\documentclass[]{revtex4-2}
\usepackage{flushend}
\usepackage{graphicx}
\usepackage{bm}
\usepackage{hyperref}
\usepackage[dvipsnames]{xcolor}
\usepackage{booktabs}
\usepackage{multirow}
\usepackage{xcolor}

\usepackage{enumitem}

\begin{document}

\title{Transfer Learning for Deep Learning-based Prediction\\of Lattice Thermal Conductivity}

\author{L. Klochko}
\affiliation{%
  Université de Lorraine, LORIA, Nancy F-54000, France
}%
\author{M. d’Aquin}%

\affiliation{%
  Université de Lorraine, LORIA, Nancy F-54000, France
}%
\author{A. Togo}
\affiliation{Center for Basic Research on Materials (CBRM), National Institute for Materials Science (NIMS), Tsukuba, Ibaraki 305-0047, Japan}

\author{L. Chaput}
\affiliation{
 Université de Lorraine, LEMTA, Nancy F-54000, France \\
 Institut Universitaire de France, 1 rue Descartes, 75231 Paris, France
}%

\begin{abstract}

Machine learning promises to accelerate the material discovery by enabling high-throughput prediction of desirable macro-properties from atomic-level descriptors or structures. However, the limited data available about precise values of these properties have been a barrier, leading to predictive models with limited precision or the ability to generalize. This is particularly true of lattice thermal conductivity (LTC): existing datasets of precise (ab initio, DFT-based) computed values are limited to a few dozen materials with little variability. Based on such datasets, we study the impact of transfer learning on both the precision and generalizability of a deep learning model (ParAIsite). We start from an existing model (MEGNet~\cite{Chen2019}) and show that improvements are obtained by fine-tuning a pre-trained version on different tasks. Interestingly, we also show that a much greater improvement is obtained when first fine-tuning it on a large datasets of low-quality approximations of LTC (based on the AGL model) and then applying a second phase of fine-tuning with our high-quality, smaller-scale datasets. The promising results obtained pave the way not only towards a greater ability to explore large databases in search of low thermal conductivity materials but also to methods enabling increasingly precise predictions in areas where quality data are rare.

\end{abstract}


\maketitle


\section{Introduction}

Machine learning models have advanced material research in several fields, including various domains of physics~\cite{Karniadakis2021,Borg2023,AsensioRamos2023,He2023}, quantum chemistry~\cite{Mater2019,Vinod2024}, drug discovery~\cite{Sarkar2023}, and cancer studies~\cite{Zhang2023,Swanson2023,Yaqoob2023}. 
In particular, recent studies have proposed various models to predict the physical properties of materials~\cite{Ihalage2022,Suzuki2022,Chen2019}. These models utilize diverse datasets, input data, and tuned neural network designs for specific purposes. However, the applicability of those models, i.e. their efficacy when applied on large databases, remains limited by their precision and generalizability, which are in turn dependent on the quality and size of the available training data. In other words, unsurprisingly, machine learning models trained on small databases of similar materials do not perform well on large databases of diverse materials~\cite{Zhang2018}.

Predicting the low lattice thermal conductivity (LTC) of crystal compounds on the basis of their structure and physical properties is one of those tasks that are made challenging by the lack of quality data. However, it is critical, as the ability to identify low-LTC materials on a scale could have profound implications for the design and optimization of materials in various applications, from electronics to energy storage~\cite{Qian2023-at,Huang2023}. The difficulty of this problem lies in the complexity of the relationship between the structure of a material and its thermal properties. Although large data sets on material properties are available through databases such as AFLOW~\cite{Calderon2015}, OQMD~\cite{Kirklin2015}, Materials Project~\cite{Jain2013}, and JARVIS~\cite{Choudhary2020} available data on LTC is either based on approximate models (such as AGL~\cite{Blanco2004,Toher2014}) and therefore not usable to build precise machine learning-based predictive models, or rely on ab initio, DFT-based computation which are too expensive to run at large scales and are therefore of very small in sizes.

In this study, we apply a two-stage transfer learning methodology as a way to take advantage of both types of dataset to achieve greater levels of precision and generalizability in deep learning models for predicting LTC. Transfer learning as applied in this article is a process in which a model trained on a first task is reused as a starting point for training on a second similar task. The idea is that initial patterns can be learned in larger, relevant data, which will bootstrap the learning in smaller, more targeted data. The aim is to show the benefits of relying on an existing model having demonstrated good performance in predicting a different property than LTC in transfer learning, but also how this can be pushed further in a second stage of transfer learning, by using the larger, low quality datasets for LTC to pre-train models for predicting the smaller, high quality datasets. 

\section{Methodology}

Transfer learning~\cite{surveyTC} involves the use of a machine learning model that has been pre-trained on a large dataset and subsequently fine-tuning it on a smaller domain-specific dataset. This strategy is particularly effective when working with limited data, as it allows us to capitalize on the knowledge gained from broader datasets, as demonstrated in multiple applications for material properties~\cite{Kong2021,surveyTC2}. Here, we describe the datasets, model, and the process used to carry out our two-stage transfer learning process.

\subsection{Data}

As mentioned above, the accuracy and reliability of any machine learning model is highly dependent on the quality of the data used for training and validation. In this work, we rely on four datasets, each contributing to our analysis. The first two are derived from calculations of the first principle of anharmonic lattice dynamics~\cite{Seko2015,Togo2015}, looking at a small set of materials with specific structures. The third is a combination of the first two, used to obtain a slightly larger and slightly more diverse dataset. Finally, the fourth can be seen to represent a large dataset that contains low-precision values for LTC.
\begin{description}
    \item[Dataset1] This dataset contains 96 materials from~\cite{Seko2015} in the rocksalt, zincblende, and wurtzite structures that could be unambiguously identified in the Material Project Database. The LTC values are obtained using the phono3py software package~\cite{phonopy-phono3py-JPCM,PhysRevLett.110.265506} using the YAML files available through the PhononDB (\url{https://github.com/atztogo/phonondb})repository. Obtaining prediction with low deviation from those values is the central motivation for this work.
    \item[Dataset2] This dataset includes thermal conductivity data for 143   half-Heusler compounds, as reported in~\cite{Miyazaki2021}. This dataset adds another set of materials, while being itself significantly more specific than the previous one: it focuses only on one specific structure, for which the range of LTC values is significantly narrower ([2.17, 34.51] W/mK, compared to [0.51,1769.00] W/mK in Dataset1). 
    \item[MIX dataset] This dataset is a combination of the two datasets mentioned above, integrating the properties of Dataset1 and Dataset2 to enhance the diversity and scope of our model training. 
    \item[AFLOW AGL dataset] This dataset contains 5578 materials obtained from the AFLOW-LIB repository~\cite{Calderon2015} together with their corresponding thermal conductivity obtained with the use of a quasi-harmonic Debye-Grüneisen model~\cite{Blanco2004,Toher2014}.
\end{description}

In addition, for all cases, logarithmic scaling is applied to all values of LTC, followed by standardization using the parameters of the corresponding dataset. For validation purposes, each data set is split 9 times, keeping 80\% for training. In other words, each dataset is associated with 9 different randomly selected validation sets that represent each 20\% of the total data set that is not used for training. Any result shown later in this article is measured as an average over those 9 validation sets and the corresponding models for a given dataset.

\subsection{Model}
\label{subsec:Selection of the Pre-Trained model}

\begin{figure}[h!]
\includegraphics[scale=0.17]{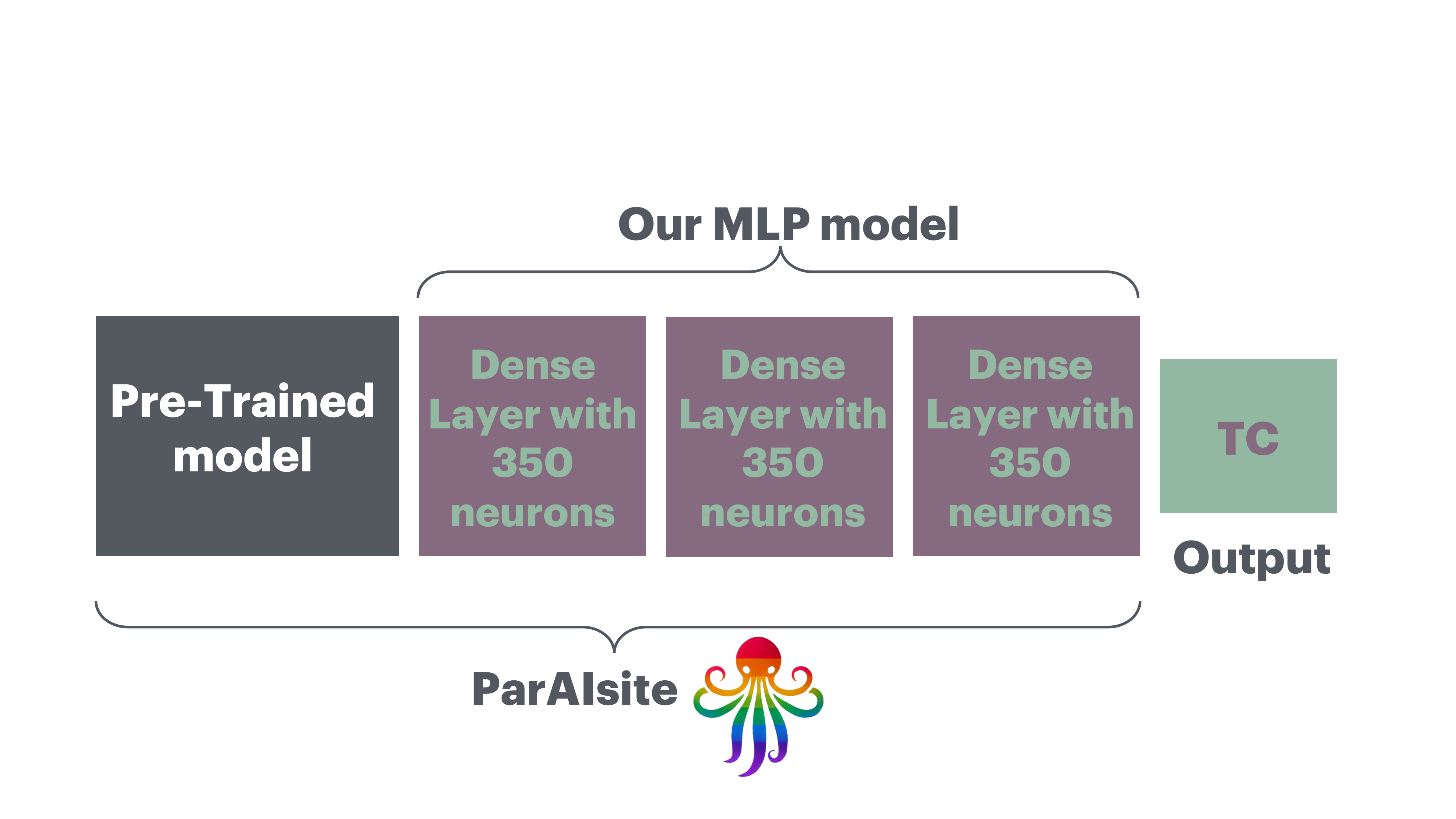}
\caption{\label{fig:paraisite_arch}The ParAIsite model architecture. Each dense layer contains 350 neurons. The output property is the thermal conductivity (TC).}
\end{figure}

Our model for predicting LTC from the properties and structure of materials (called ParAIsite) is based, as shown in Figure~\ref{fig:paraisite_arch}, on the addition of a multilayer perceptron (MLP, a dense, feed-forward, fully connected layer) on top of an existing, pre-trained model. In line with our transfer learning approach, the idea here is to use an existing model (the pre-trained model) that has already shown its ability to predict properties of materials as a foundation to be adapted for the task of LTC prediction. More concretely, ParAIsite is based on connecting the last hidden layer of the pre-trained model to the input of the added MLP.  

A first step, therefore, for establishing this model is the selection of the most appropriate pre-trained model from the top-performing models cataloged on MatBench~\cite{Dunn2020}. 
We only considered models based on an unambiguous identification of the materials as input and on a representation of their structures (in the form of  crystallographic information files, CIFs). An initial set of tests were carried out with Dataset1 using the model of Figure~\ref{fig:paraisite_arch} with each candidate pre-trained model to validate the model's performance and ascertain its suitability for the specific challenges associated with predicting the thermal conductivity in crystal compounds. 

\begin{figure}[h!]
\includegraphics[scale=0.45]{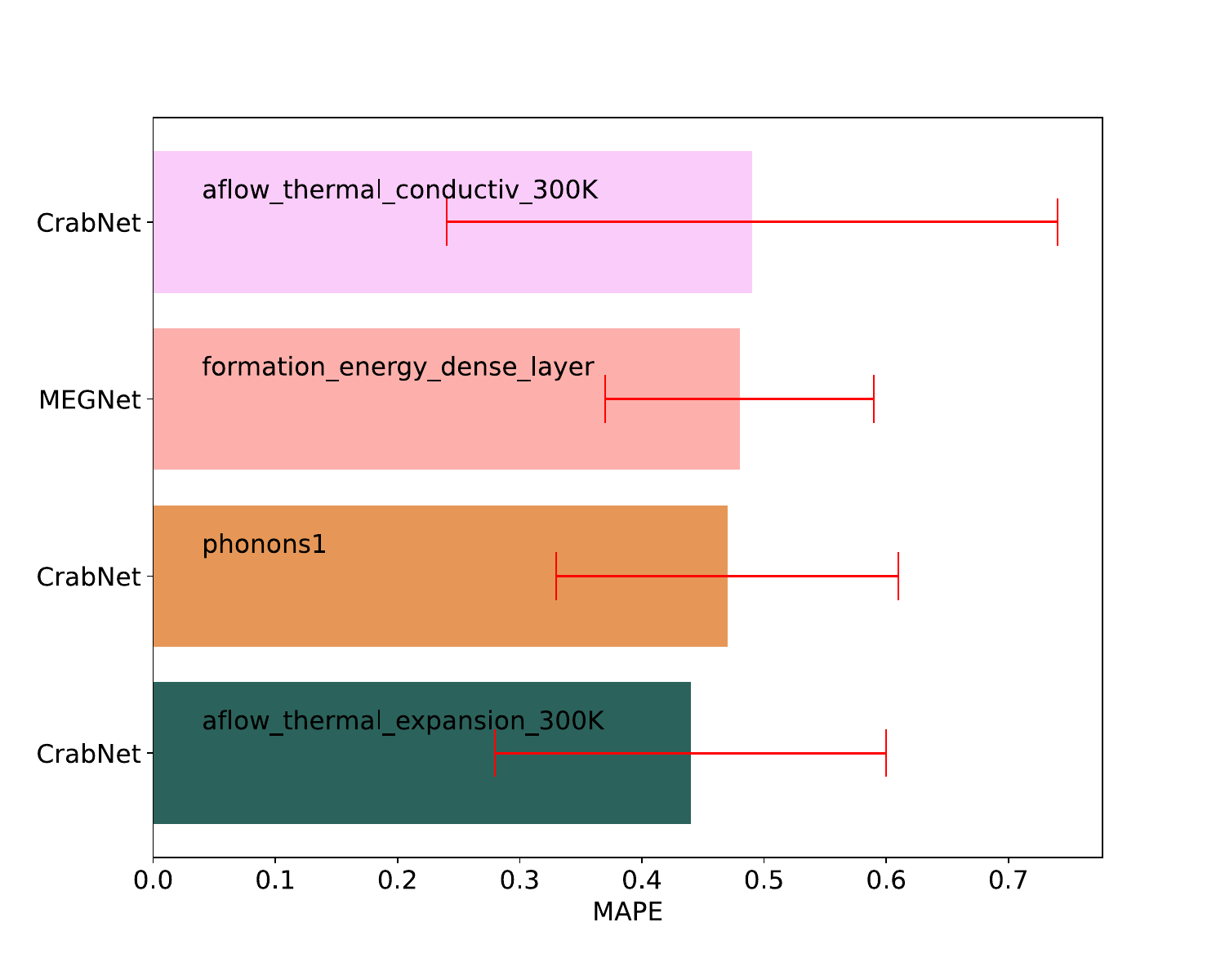}
\caption{\label{fig:mape_mlp_test}Results of validation tests of ParAIsite when using top-performing models cataloged on MatBench~\cite{Dunn2020} as pre-trained models. As one can see, the MEGNet model~\cite{Chen2019} shows better stability compared to the CrabNet model~\cite{Wang2021crabnet} on Dataset1.  Here, the name of the datasets on which model was pre-trained are  indicated inside the box, and the error variablity are shown in red lines. }
\end{figure}

Following validation (see Fig.~\ref{fig:mape_mlp_test}), we chose the model that combined the best performance (measured by the mean absolute percentage error, MAPE) and was most consistent with the features of our dataset. Despite the better average performance of the CrabNet model~\cite{Wang2021crabnet} on Dataset1, results over multiple runs showed a lot of variability, demonstrating that this model was too unstable to be used.  That result led us to use the graph-based neural network MEGNet model~\cite{Chen2019}, which was pre-trained on the formation energy data of 62,315 compounds from the Materials Project Database as of the 2018.6.1 version.  By fine-tuning MEGNet on our thermal conductivity data, we aim to improve the prediction accuracy and better understand the thermal properties of the crystal compounds in our data set.

\subsection{Model training}

In order to solve the complexity of predicting LTC, we adopted a three-step method to systematically improve model performance and evaluate the impact  of transfer learning (TL) between the steps. In summary, those steps move from no transfer learning at all, to two levels of pre-training/transfer learning being applied, allowing us to evaluate how pre-trained models and fine-tuning affect the model's ability to generalize across datasets.

\begin{description}

\item[Step~1 (no pretraining)] In this step, we train and test ParAIsite on our four datasets using an uninitialized MEGNet (and MLP) model. The training is therefore done from scratch (with random initial weights), without any form of transfer learning.
   
\item[Step~2 (using a pre-trained MEGNet)] In this step, we train and test ParAIsite on our four datasets using a pre-trained MEGNet, that is, where MEGNet is initialized with the weights obtained through the training carried out by its authors on the task of predicting formation energy.
   
\item[Step~3 (transfer-learning over fine-tuned AFLOW model)] In this step, we train and test ParAIsite on Dataset1, Dataset2 and MIX, taking as a starting point the best model after fine-tuning using the AFLOW~AGL dataset. In other words, we apply a second round of fine-tuning, over the one already done for MEGNet, and our own on AFLOW~AGL. By pretraining the whole model further on a larger, but lower quality dataset, we anticipate that training on the three other, more precise datasets will converge to better performances.
 
\end{description}

At each step, for each model, training is performed for 300 epochs with a fixed random seed of 42, ensuring reproducible results. 
As already mentioned, each of these steps is repeated 9 times to ensure statistical robustness, and the averaged results for validation loss across all steps are presented in Figs.~\ref{fig:loss_all_steps_L96}--\ref{fig:loss_all_steps_MIX}. For all training steps, we use MAPE as the loss function, having applied normalisation and scaling. 

\section{Results and discussion}


A first straightforward conclusion that can be drawn from training and cross-validating in each step with the considered datasets is that the double steps of transfer learning is having a significant effect on the performance of the model on Dataset1, and generally improve its capacity to generalize if the considered dataset is varied enough. Indeed, in Table~\ref{tab:1} we show the average and standard deviation (over the 9 runs) of MAPE of models trained on the training subsets of each dataset (rows) and tested on the validation subsets of each dataset (columns).\\

\begin{table}[h!]
\centering
\begin{tabular}{|l|c|c|c|c|}
\toprule
\textbf{Train on \textbackslash Test on} & \textbf{Dataset1} & \textbf{Dataset2} & \textbf{MIX} & \textbf{AFLOW} \\
\midrule
\textbf{Step 1:} & & & & \\
Dataset1   & 0.82 (0.21) & 2.53 (2.17) & 1.74 (1.26) & 2.16 (1.26) \\
Dataset2 & 0.51 (0.09) & 0.42 (0.07) & 0.43 (0.05) & 0.48 (0.05) \\
MIX   & 0.69 (0.15) & 0.77 (0.15) & 0.83 (0.07) & 0.97 (0.27) \\
AFLOW & 0.55 (0.34) & 1.18 (0.27) & 0.93 (0.27) & 0.62 (0.33) \\
\midrule
\textbf{Step 2:} & & & & \\
Dataset1   & 0.76 (0.29) & 3.09 (2.40) & 2.07 (1.40) & 2.32 (1.94) \\
Dataset2 & 0.55 (0.14) & 0.42 (0.06) & 0.44 (0.04) & 0.52 (0.10) \\
MIX   & 0.66 (0.14) & 0.75 (0.16) & 0.81 (0.11) & 1.10 (0.47) \\
AFLOW & 0.44 (0.14) & 1.27 (0.48) & 0.94 (0.30) & 0.50 (0.03) \\
\midrule
\textbf{Step 3:} & & & & \\
Dataset1   & 0.34 (0.15) & 1.26 (0.44) & 0.87 (0.27) & 0.57 (0.10) \\
Dataset2 & 0.55 (0.10) & 0.78 (0.20) & 0.63 (0.18) & 0.62 (0.07) \\
MIX   & 0.37 (0.14) & 0.78 (0.31) & 0.69 (0.19) & 0.59 (0.06) \\
\bottomrule
\end{tabular}
\caption{\label{tab:1}Validation error (MAPE) across training steps and datasets.}
\end{table}

Taking as example Dataset1, which is considered particularly challenging, we can see that \textcolor{red}{with} no transfer learning, models reach on average a 82\% error on average when tested on Dataset1's validation subsets. The error, training and testing again on the relevant subsets of Dataset1 falls to 76\% when using the original pre-trained MEGNet model (Step~1) falls to an average of 76\%, and falls significantly further, to 34\% at Step~2. In other words, transfer learning has had a significant effect, especially taking as starting point a model that has already been fine-tuned for a similar task (i.e. predicting approximated LTC through the AFLOW dataset).

\begin{figure}[h!]
\includegraphics[scale=0.45]{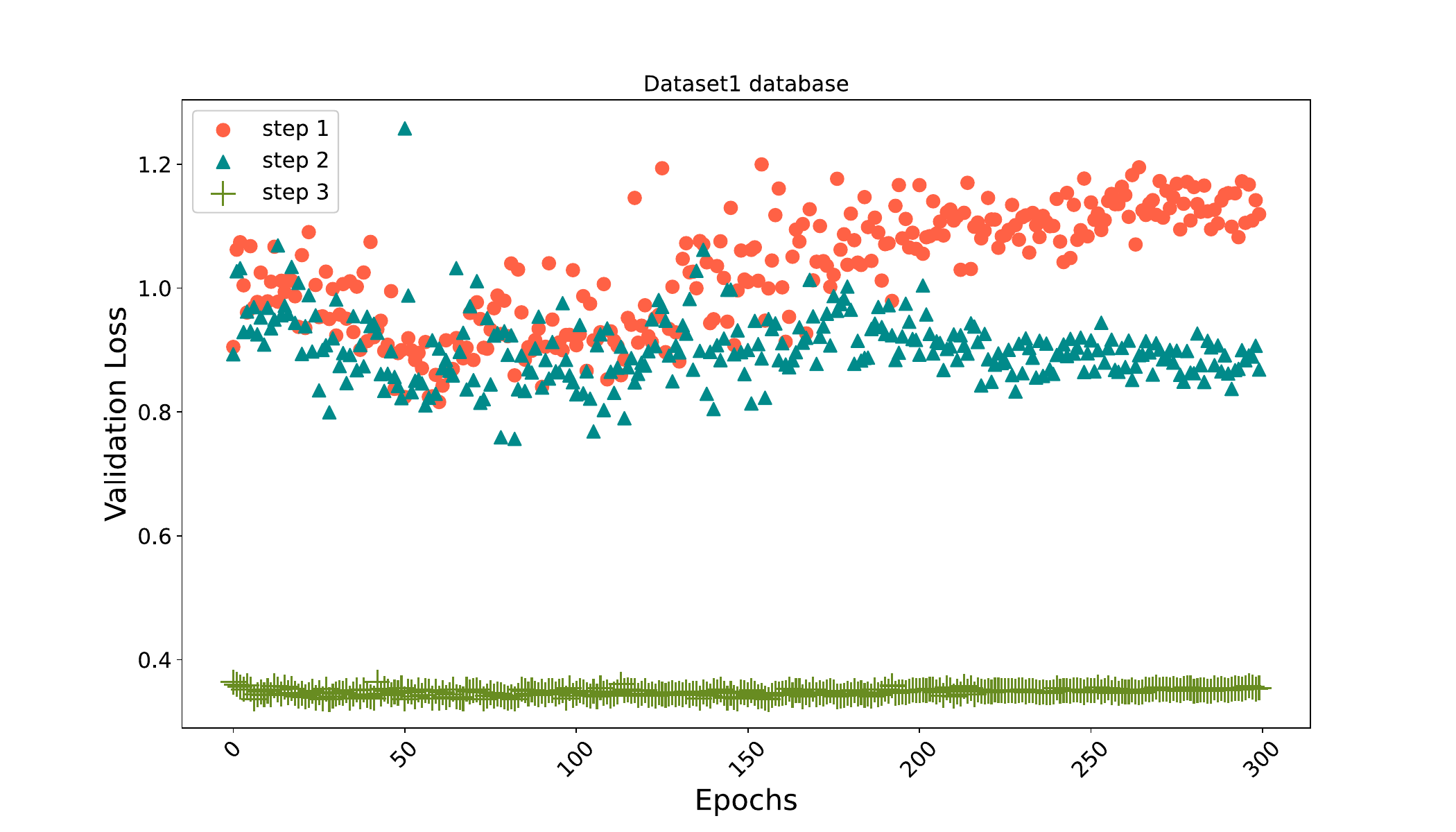}
\caption{\label{fig:loss_all_steps_L96}  Validation loss (MAPE) for models trained and tested on Dataset1 across training epochs.}
\end{figure}

This effect of the two steps of transfer learning is particularly visible in Figure~\ref{fig:loss_all_steps_L96}, which shows the evolution of the average MAPE (over 9 runs) in the validation subsets of Dataset1 during the training iterations (epochs). Comparing this evolution between Step~1 and Step~2 shows that starting from a relevant pre-trained model, even if made for a different task, enabled the model to converge faster to slightly lower values of MAPE. We can also see that the MAPE on the validation subsets does not rise up in Step~2 as much as it did during the Step~1 training process, showing that the model was less prone to overfitting in this case. Looking at the chart for Step~3, we can see a significantly different behavior, with the MAPE value converging very quickly to much lower values.\\

\begin{figure}[h!]
\includegraphics[scale=0.45]{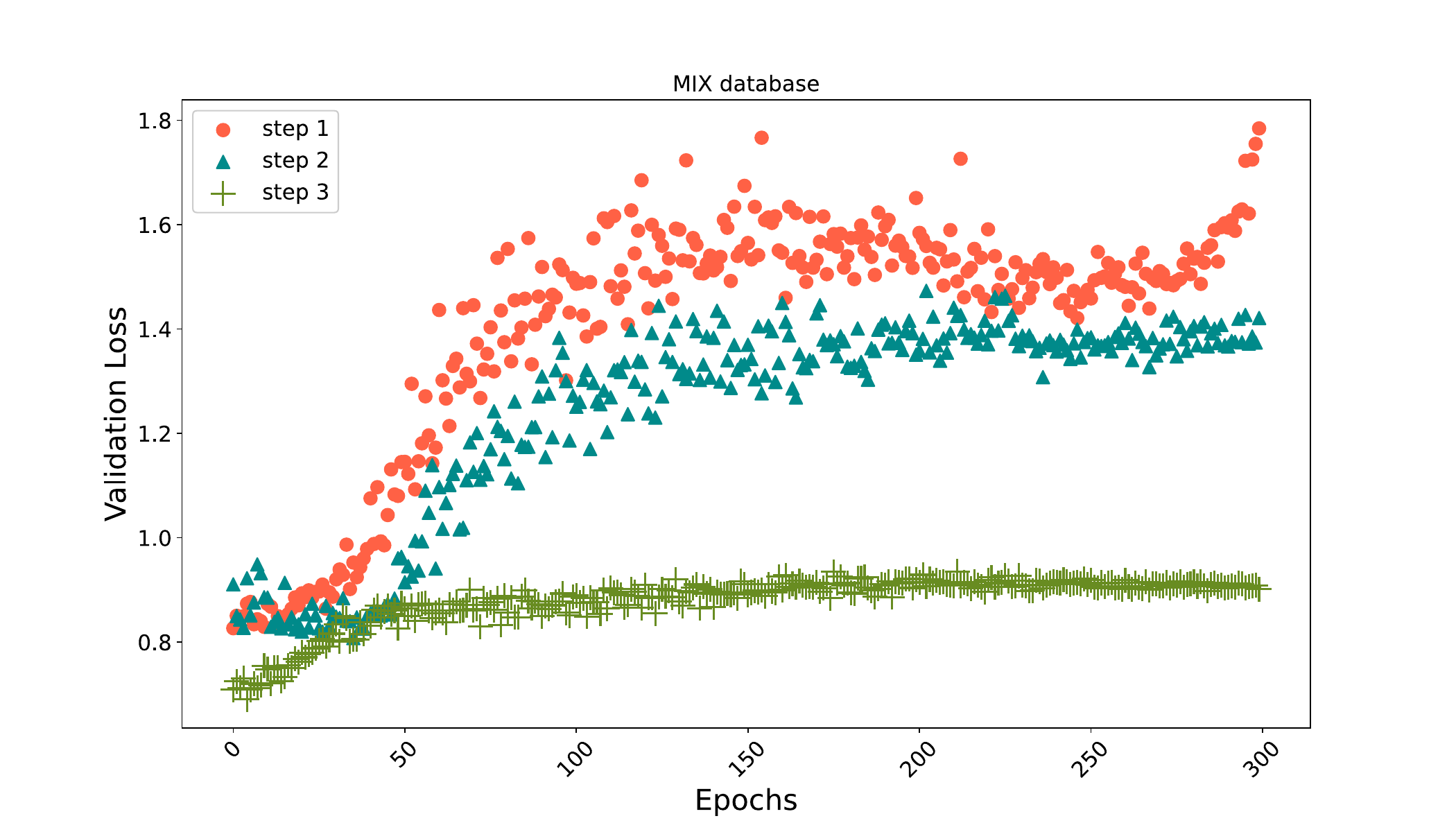}
\caption{\label{fig:loss_all_steps_MIX}  Validation loss (MAPE) for models trained and tested on MIX across training epochs.}
\end{figure}

Similar conclusions can be drawn on the MIX dataset, even if those are less strong: The models trained on their training subsets and tested on their validation subsets reach 83\%, 81\% and 69\% in steps~1,~2 and~3 respectively. As seen in Figure~\ref{fig:loss_all_steps_MIX}, MAPE also reaches lower values in Step~2 compared to Step~2 and significantly lower ones yet in Step~3. However, in this case, overfitting appears early in the training process (around epoch~50) in all steps. This is probably explained by the fact that the MIX dataset combines Dataset1, which is well predicted through the transfer learning process, and Dataset2 which, as we discuss below, did not work as well.\\

\begin{figure}[h!]
\includegraphics[scale=0.45]{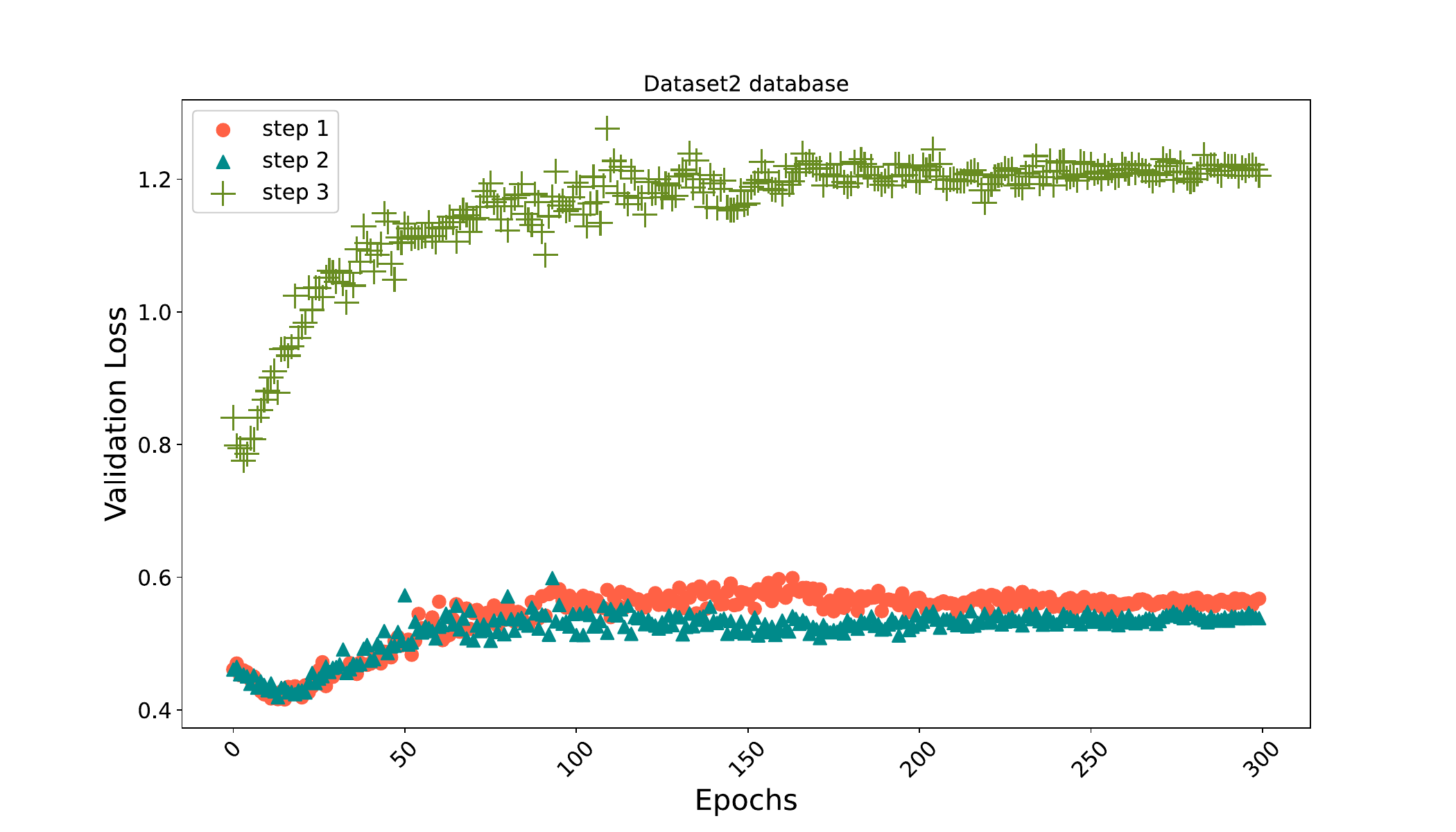}
\caption{\label{fig:loss_all_steps_HH143} Validation loss (MAPE) for models trained and tested on Dataset2 across training epochs. }
\end{figure}

Regarding Dataset2, the results shown in Table~\ref{tab:1}, the results obtained are against the ones obtained for the other two datasets. In fact, when training and testing on subsets of Dataset2, the best average MAPE obtained where 42\%, 42\% and 78\% in steps~1,~2 and~3 respectively. In other words, the third step, the double pre-training, significantly worsens the results obtained. This can be explained by the fact that this dataset contains a very restricted range of LTCs compared to the others, in particular AFLOW. In other words, having learned in Step~2 to predict a wide range of LTC values through the AFLOW dataset, the model did not adapt well to the very specific set of materials in Dataset2. We can observe this issue in Figure~\ref{fig:loss_all_steps_HH143} where, in Step~3, the training starts with a high error rate and is overfit almost immediately.\\

The issue mentioned above, of the lack of variety in Dataset2, is part of the motivation for integrating the MIX dataset as well. One final conclusion that can be drawn from the results shown in Table~\ref{tab:1} is that, in accordance with our assumption, a more varied dataset (MIX) tends to generalize better, but also that the double pre-training process applied here helps support this ability to generalize. This is visible in the last line of the table, which shows that models trained on the MIX dataset are better on average across all datasets (including AFLOW) than those trained on the other two datasets in the same step. In most cases, models trained on MIX or Dataset1 achieved better results when tested with other datasets at Step~3 than at other steps.

\section{Conclusion}

In this work, we tested using multiple steps of transfer learning (pre-training and fine-tuning) a machine learning model that we developed to predict the thermal conductivity of crystal compounds. Despite the availability of large datasets for general material properties, databases specifically focused on thermal conductivity are limited, which introduced a challenge to our task. For this reason, the model was trained and validated on three different high-precision datasets, one being a mixture of the other two. We also used a less precise, but larger volume dataset as part of one of the transfer learning steps.

Through a series of experiments involving different training configurations, we found that double transfer learning, which includes an additional phase of training on external data, proved to be effective in reaching not only better precision but also better generalizability and reduced overfitting. This is true for the datasets that show a broader range of values for LTC, and more diversity in the type of material included. For those, the error rate (MAPE) decreased consistently as we progressed through the steps, particularly in Step~3. This indicates that transfer learning, when applied judiciously, can enhance model performance on small datasets.

However, for our less varied dataset, transfer learning had the opposite effect. Double TL caused the model to rapidly overfit, as the specific nature of the dataset was incompatible with the broader generalization achieved through additional external training. In other words, transfer learning was most effective when the dataset was representative of a wide range of materials. In contrast, when dealing with highly specialized datasets, additional training phases may introduce confusion rather than improvement.

To provide concrete validation of the best performing models, we applied them to obtain predictions for stable materials in the Material Project Database. LTC for $(BaSbO_3)_2$ (mp-9127) was then calculated using a robust ab initio calculation, as our models consistently found that it has a relatively low thermal conductivity. The result of the computation (7.1 W / m * K) was on the same order of magnitude as the predictions of our models (1.23  W/m*K). This agreement underscores the ability of the model to capture critical trends in LTC prediction, even for datasets it was not trained directly on.

In summary, while double transfer learning shows great promise in improving model accuracy and generalization, its success heavily depends on the dataset's diversity and scope. The results obtained are very promising but also demonstrate that the choice of dataset and training approach is crucial when predicting thermal conductivity. A greater availability of a broader range of datasets of LTC, whether of high precision for training or of approximate precision for pre-training, is therefore expected to enable us to reach better results in the future.

\section{Code and data availability}

The code and data that are required to reproduce the results of paper or re-utilizing for their own purposes are shown on \url{https://github.com/liudakl/ParAIsite.git}

\begin{acknowledgments}
The authors acknowledge Lorraine Université d’Excellence (\url{https://www.univ-lorraine.fr/lue/}) for the financial support of Liudmyla Klochko to work on this project.  
\end{acknowledgments}

\section{Author contributions}
All authors discussed the results and contributed equally to the manuscript. L. Klochko developed and carried out the implementation of the model and wrote the first versions of the manuscript. M. d'Aquin provided the domain expertise in deep learning and transfer learning necessary for this work and revised the manuscript. L. Chaput provided the necessary domain expertise in crystallography and thermophysical properties of material and revised the manuscript. A. Togo developed the computer codes used to compute the thermal conductivity from the first principle.

\section{Competing interests }
There is no competing interests to declare. 

\bibliography{biblio}

\providecommand{\noopsort}[1]{}\providecommand{\singleletter}[1]{#1}%
\begin{thebibliography}{32}%
\makeatletter
\providecommand \@ifxundefined [1]{%
 \@ifx{#1\undefined}
}%
\providecommand \@ifnum [1]{%
 \ifnum #1\expandafter \@firstoftwo
 \else \expandafter \@secondoftwo
 \fi
}%
\providecommand \@ifx [1]{%
 \ifx #1\expandafter \@firstoftwo
 \else \expandafter \@secondoftwo
 \fi
}%
\providecommand \natexlab [1]{#1}%
\providecommand \enquote  [1]{``#1''}%
\providecommand \bibnamefont  [1]{#1}%
\providecommand \bibfnamefont [1]{#1}%
\providecommand \citenamefont [1]{#1}%
\providecommand \href@noop [0]{\@secondoftwo}%
\providecommand \href [0]{\begingroup \@sanitize@url \@href}%
\providecommand \@href[1]{\@@startlink{#1}\@@href}%
\providecommand \@@href[1]{\endgroup#1\@@endlink}%
\providecommand \@sanitize@url [0]{\catcode `\\12\catcode `\$12\catcode `\&12\catcode `\#12\catcode `\^12\catcode `\_12\catcode `\%12\relax}%
\providecommand \@@startlink[1]{}%
\providecommand \@@endlink[0]{}%
\providecommand \url  [0]{\begingroup\@sanitize@url \@url }%
\providecommand \@url [1]{\endgroup\@href {#1}{\urlprefix }}%
\providecommand \urlprefix  [0]{URL }%
\providecommand \Eprint [0]{\href }%
\providecommand \doibase [0]{https://doi.org/}%
\providecommand \selectlanguage [0]{\@gobble}%
\providecommand \bibinfo  [0]{\@secondoftwo}%
\providecommand \bibfield  [0]{\@secondoftwo}%
\providecommand \translation [1]{[#1]}%
\providecommand \BibitemOpen [0]{}%
\providecommand \bibitemStop [0]{}%
\providecommand \bibitemNoStop [0]{.\EOS\space}%
\providecommand \EOS [0]{\spacefactor3000\relax}%
\providecommand \BibitemShut  [1]{\csname bibitem#1\endcsname}%
\let\auto@bib@innerbib\@empty
\bibitem [{\citenamefont {Chen}\ \emph {et~al.}(2019)\citenamefont {Chen}, \citenamefont {Ye}, \citenamefont {Zuo}, \citenamefont {Zheng},\ and\ \citenamefont {Ong}}]{Chen2019}%
  \BibitemOpen
  \bibfield  {author} {\bibinfo {author} {\bibfnamefont {C.}~\bibnamefont {Chen}}, \bibinfo {author} {\bibfnamefont {W.}~\bibnamefont {Ye}}, \bibinfo {author} {\bibfnamefont {Y.}~\bibnamefont {Zuo}}, \bibinfo {author} {\bibfnamefont {C.}~\bibnamefont {Zheng}},\ and\ \bibinfo {author} {\bibfnamefont {S.~P.}\ \bibnamefont {Ong}},\ }\bibfield  {title} {\bibinfo {title} {Graph networks as a universal machine learning framework for molecules and crystals},\ }\href {https://doi.org/10.1021/acs.chemmater.9b01294} {\bibfield  {journal} {\bibinfo  {journal} {Chemistry of Materials}\ }\textbf {\bibinfo {volume} {31}},\ \bibinfo {pages} {3564–3572} (\bibinfo {year} {2019})}\BibitemShut {NoStop}%
\bibitem [{\citenamefont {Karniadakis}\ \emph {et~al.}(2021)\citenamefont {Karniadakis}, \citenamefont {Kevrekidis}, \citenamefont {Lu}, \citenamefont {Perdikaris}, \citenamefont {Wang},\ and\ \citenamefont {Yang}}]{Karniadakis2021}%
  \BibitemOpen
  \bibfield  {author} {\bibinfo {author} {\bibfnamefont {G.~E.}\ \bibnamefont {Karniadakis}}, \bibinfo {author} {\bibfnamefont {I.~G.}\ \bibnamefont {Kevrekidis}}, \bibinfo {author} {\bibfnamefont {L.}~\bibnamefont {Lu}}, \bibinfo {author} {\bibfnamefont {P.}~\bibnamefont {Perdikaris}}, \bibinfo {author} {\bibfnamefont {S.}~\bibnamefont {Wang}},\ and\ \bibinfo {author} {\bibfnamefont {L.}~\bibnamefont {Yang}},\ }\bibfield  {title} {\bibinfo {title} {Physics-informed machine learning},\ }\href {https://doi.org/10.1038/s42254-021-00314-5} {\bibfield  {journal} {\bibinfo  {journal} {Nature Reviews Physics}\ }\textbf {\bibinfo {volume} {3}},\ \bibinfo {pages} {422–440} (\bibinfo {year} {2021})}\BibitemShut {NoStop}%
\bibitem [{\citenamefont {Borg}\ \emph {et~al.}(2023)\citenamefont {Borg}, \citenamefont {Muckley}, \citenamefont {Nyby}, \citenamefont {Saal}, \citenamefont {Ward}, \citenamefont {Mehta},\ and\ \citenamefont {Meredig}}]{Borg2023}%
  \BibitemOpen
  \bibfield  {author} {\bibinfo {author} {\bibfnamefont {C.~K.~H.}\ \bibnamefont {Borg}}, \bibinfo {author} {\bibfnamefont {E.~S.}\ \bibnamefont {Muckley}}, \bibinfo {author} {\bibfnamefont {C.}~\bibnamefont {Nyby}}, \bibinfo {author} {\bibfnamefont {J.~E.}\ \bibnamefont {Saal}}, \bibinfo {author} {\bibfnamefont {L.}~\bibnamefont {Ward}}, \bibinfo {author} {\bibfnamefont {A.}~\bibnamefont {Mehta}},\ and\ \bibinfo {author} {\bibfnamefont {B.}~\bibnamefont {Meredig}},\ }\bibfield  {title} {\bibinfo {title} {Quantifying the performance of machine learning models in materials discovery},\ }\href {https://doi.org/10.1039/d2dd00113f} {\bibfield  {journal} {\bibinfo  {journal} {Digital Discovery}\ }\textbf {\bibinfo {volume} {2}},\ \bibinfo {pages} {327–338} (\bibinfo {year} {2023})}\BibitemShut {NoStop}%
\bibitem [{\citenamefont {Asensio~Ramos}\ \emph {et~al.}(2023)\citenamefont {Asensio~Ramos}, \citenamefont {Cheung}, \citenamefont {Chifu},\ and\ \citenamefont {Gafeira}}]{AsensioRamos2023}%
  \BibitemOpen
  \bibfield  {author} {\bibinfo {author} {\bibfnamefont {A.}~\bibnamefont {Asensio~Ramos}}, \bibinfo {author} {\bibfnamefont {M.~C.~M.}\ \bibnamefont {Cheung}}, \bibinfo {author} {\bibfnamefont {I.}~\bibnamefont {Chifu}},\ and\ \bibinfo {author} {\bibfnamefont {R.}~\bibnamefont {Gafeira}},\ }\bibfield  {title} {\bibinfo {title} {Machine learning in solar physics},\ }\bibfield  {journal} {\bibinfo  {journal} {Living Reviews in Solar Physics}\ }\textbf {\bibinfo {volume} {20}},\ \href {https://doi.org/10.1007/s41116-023-00038-x} {10.1007/s41116-023-00038-x} (\bibinfo {year} {2023})\BibitemShut {NoStop}%
\bibitem [{\citenamefont {He}\ \emph {et~al.}(2023)\citenamefont {He}, \citenamefont {Ma}, \citenamefont {Pang}, \citenamefont {Song},\ and\ \citenamefont {Zhou}}]{He2023}%
  \BibitemOpen
  \bibfield  {author} {\bibinfo {author} {\bibfnamefont {W.-B.}\ \bibnamefont {He}}, \bibinfo {author} {\bibfnamefont {Y.-G.}\ \bibnamefont {Ma}}, \bibinfo {author} {\bibfnamefont {L.-G.}\ \bibnamefont {Pang}}, \bibinfo {author} {\bibfnamefont {H.-C.}\ \bibnamefont {Song}},\ and\ \bibinfo {author} {\bibfnamefont {K.}~\bibnamefont {Zhou}},\ }\bibfield  {title} {\bibinfo {title} {High-energy nuclear physics meets machine learning},\ }\bibfield  {journal} {\bibinfo  {journal} {Nuclear Science and Techniques}\ }\textbf {\bibinfo {volume} {34}},\ \href {https://doi.org/10.1007/s41365-023-01233-z} {10.1007/s41365-023-01233-z} (\bibinfo {year} {2023})\BibitemShut {NoStop}%
\bibitem [{\citenamefont {Mater}\ and\ \citenamefont {Coote}(2019)}]{Mater2019}%
  \BibitemOpen
  \bibfield  {author} {\bibinfo {author} {\bibfnamefont {A.~C.}\ \bibnamefont {Mater}}\ and\ \bibinfo {author} {\bibfnamefont {M.~L.}\ \bibnamefont {Coote}},\ }\bibfield  {title} {\bibinfo {title} {Deep learning in chemistry},\ }\href {https://doi.org/10.1021/acs.jcim.9b00266} {\bibfield  {journal} {\bibinfo  {journal} {Journal of Chemical Information and Modeling}\ }\textbf {\bibinfo {volume} {59}},\ \bibinfo {pages} {2545–2559} (\bibinfo {year} {2019})}\BibitemShut {NoStop}%
\bibitem [{\citenamefont {Vinod}\ \emph {et~al.}(2024)\citenamefont {Vinod}, \citenamefont {Kleinekath\"{o}fer},\ and\ \citenamefont {Zaspel}}]{Vinod2024}%
  \BibitemOpen
  \bibfield  {author} {\bibinfo {author} {\bibfnamefont {V.}~\bibnamefont {Vinod}}, \bibinfo {author} {\bibfnamefont {U.}~\bibnamefont {Kleinekath\"{o}fer}},\ and\ \bibinfo {author} {\bibfnamefont {P.}~\bibnamefont {Zaspel}},\ }\bibfield  {title} {\bibinfo {title} {Optimized multifidelity machine learning for quantum chemistry},\ }\href {https://doi.org/10.1088/2632-2153/ad2cef} {\bibfield  {journal} {\bibinfo  {journal} {Machine Learning: Science and Technology}\ }\textbf {\bibinfo {volume} {5}},\ \bibinfo {pages} {015054} (\bibinfo {year} {2024})}\BibitemShut {NoStop}%
\bibitem [{\citenamefont {Sarkar}\ \emph {et~al.}(2023)\citenamefont {Sarkar}, \citenamefont {Das}, \citenamefont {Rawat}, \citenamefont {Wahlang}, \citenamefont {Nongpiur}, \citenamefont {Tiewsoh}, \citenamefont {Lyngdoh}, \citenamefont {Das}, \citenamefont {Bidarolli},\ and\ \citenamefont {Sony}}]{Sarkar2023}%
  \BibitemOpen
  \bibfield  {author} {\bibinfo {author} {\bibfnamefont {C.}~\bibnamefont {Sarkar}}, \bibinfo {author} {\bibfnamefont {B.}~\bibnamefont {Das}}, \bibinfo {author} {\bibfnamefont {V.~S.}\ \bibnamefont {Rawat}}, \bibinfo {author} {\bibfnamefont {J.~B.}\ \bibnamefont {Wahlang}}, \bibinfo {author} {\bibfnamefont {A.}~\bibnamefont {Nongpiur}}, \bibinfo {author} {\bibfnamefont {I.}~\bibnamefont {Tiewsoh}}, \bibinfo {author} {\bibfnamefont {N.~M.}\ \bibnamefont {Lyngdoh}}, \bibinfo {author} {\bibfnamefont {D.}~\bibnamefont {Das}}, \bibinfo {author} {\bibfnamefont {M.}~\bibnamefont {Bidarolli}},\ and\ \bibinfo {author} {\bibfnamefont {H.~T.}\ \bibnamefont {Sony}},\ }\bibfield  {title} {\bibinfo {title} {Artificial intelligence and machine learning technology driven modern drug discovery and development},\ }\href {https://doi.org/10.3390/ijms24032026} {\bibfield  {journal} {\bibinfo  {journal} {International Journal of Molecular Sciences}\ }\textbf {\bibinfo {volume} {24}},\ \bibinfo {pages} {2026} (\bibinfo {year}
  {2023})}\BibitemShut {NoStop}%
\bibitem [{\citenamefont {Zhang}\ \emph {et~al.}(2023)\citenamefont {Zhang}, \citenamefont {Shi},\ and\ \citenamefont {Wang}}]{Zhang2023}%
  \BibitemOpen
  \bibfield  {author} {\bibinfo {author} {\bibfnamefont {B.}~\bibnamefont {Zhang}}, \bibinfo {author} {\bibfnamefont {H.}~\bibnamefont {Shi}},\ and\ \bibinfo {author} {\bibfnamefont {H.}~\bibnamefont {Wang}},\ }\bibfield  {title} {\bibinfo {title} {Machine learning and ai in cancer prognosis, prediction, and treatment selection: A critical approach},\ }\href {https://doi.org/10.2147/jmdh.s410301} {\bibfield  {journal} {\bibinfo  {journal} {Journal of Multidisciplinary Healthcare}\ }\textbf {\bibinfo {volume} {Volume 16}},\ \bibinfo {pages} {1779–1791} (\bibinfo {year} {2023})}\BibitemShut {NoStop}%
\bibitem [{\citenamefont {Swanson}\ \emph {et~al.}(2023)\citenamefont {Swanson}, \citenamefont {Wu}, \citenamefont {Zhang}, \citenamefont {Alizadeh},\ and\ \citenamefont {Zou}}]{Swanson2023}%
  \BibitemOpen
  \bibfield  {author} {\bibinfo {author} {\bibfnamefont {K.}~\bibnamefont {Swanson}}, \bibinfo {author} {\bibfnamefont {E.}~\bibnamefont {Wu}}, \bibinfo {author} {\bibfnamefont {A.}~\bibnamefont {Zhang}}, \bibinfo {author} {\bibfnamefont {A.~A.}\ \bibnamefont {Alizadeh}},\ and\ \bibinfo {author} {\bibfnamefont {J.}~\bibnamefont {Zou}},\ }\bibfield  {title} {\bibinfo {title} {From patterns to patients: Advances in clinical machine learning for cancer diagnosis, prognosis, and treatment},\ }\href {https://doi.org/10.1016/j.cell.2023.01.035} {\bibfield  {journal} {\bibinfo  {journal} {Cell}\ }\textbf {\bibinfo {volume} {186}},\ \bibinfo {pages} {1772–1791} (\bibinfo {year} {2023})}\BibitemShut {NoStop}%
\bibitem [{\citenamefont {Yaqoob}\ \emph {et~al.}(2023)\citenamefont {Yaqoob}, \citenamefont {Musheer~Aziz},\ and\ \citenamefont {verma}}]{Yaqoob2023}%
  \BibitemOpen
  \bibfield  {author} {\bibinfo {author} {\bibfnamefont {A.}~\bibnamefont {Yaqoob}}, \bibinfo {author} {\bibfnamefont {R.}~\bibnamefont {Musheer~Aziz}},\ and\ \bibinfo {author} {\bibfnamefont {N.~K.}\ \bibnamefont {verma}},\ }\bibfield  {title} {\bibinfo {title} {Applications and techniques of machine learning in cancer classification: A systematic review},\ }\href {https://doi.org/10.1007/s44230-023-00041-3} {\bibfield  {journal} {\bibinfo  {journal} {Human-Centric Intelligent Systems}\ }\textbf {\bibinfo {volume} {3}},\ \bibinfo {pages} {588–615} (\bibinfo {year} {2023})}\BibitemShut {NoStop}%
\bibitem [{\citenamefont {Ihalage}\ and\ \citenamefont {Hao}(2022)}]{Ihalage2022}%
  \BibitemOpen
  \bibfield  {author} {\bibinfo {author} {\bibfnamefont {A.}~\bibnamefont {Ihalage}}\ and\ \bibinfo {author} {\bibfnamefont {Y.}~\bibnamefont {Hao}},\ }\bibfield  {title} {\bibinfo {title} {Formula graph self‐attention network for representation‐domain independent materials discovery},\ }\bibfield  {journal} {\bibinfo  {journal} {Advanced Science}\ }\textbf {\bibinfo {volume} {9}},\ \href {https://doi.org/10.1002/advs.202200164} {10.1002/advs.202200164} (\bibinfo {year} {2022})\BibitemShut {NoStop}%
\bibitem [{\citenamefont {Suzuki}\ \emph {et~al.}(2022)\citenamefont {Suzuki}, \citenamefont {Taniai}, \citenamefont {Saito}, \citenamefont {Ushiku},\ and\ \citenamefont {Ono}}]{Suzuki2022}%
  \BibitemOpen
  \bibfield  {author} {\bibinfo {author} {\bibfnamefont {Y.}~\bibnamefont {Suzuki}}, \bibinfo {author} {\bibfnamefont {T.}~\bibnamefont {Taniai}}, \bibinfo {author} {\bibfnamefont {K.}~\bibnamefont {Saito}}, \bibinfo {author} {\bibfnamefont {Y.}~\bibnamefont {Ushiku}},\ and\ \bibinfo {author} {\bibfnamefont {K.}~\bibnamefont {Ono}},\ }\bibfield  {title} {\bibinfo {title} {Self-supervised learning of materials concepts from crystal structures via deep neural networks},\ }\href {https://doi.org/10.1088/2632-2153/aca23d} {\bibfield  {journal} {\bibinfo  {journal} {Machine Learning: Science and Technology}\ }\textbf {\bibinfo {volume} {3}},\ \bibinfo {pages} {045034} (\bibinfo {year} {2022})}\BibitemShut {NoStop}%
\bibitem [{\citenamefont {Zhang}\ and\ \citenamefont {Ling}(2018)}]{Zhang2018}%
  \BibitemOpen
  \bibfield  {author} {\bibinfo {author} {\bibfnamefont {Y.}~\bibnamefont {Zhang}}\ and\ \bibinfo {author} {\bibfnamefont {C.}~\bibnamefont {Ling}},\ }\bibfield  {title} {\bibinfo {title} {A strategy to apply machine learning to small datasets in materials science},\ }\bibfield  {journal} {\bibinfo  {journal} {npj Computational Materials}\ }\textbf {\bibinfo {volume} {4}},\ \href {https://doi.org/10.1038/s41524-018-0081-z} {10.1038/s41524-018-0081-z} (\bibinfo {year} {2018})\BibitemShut {NoStop}%
\bibitem [{\citenamefont {Qian}\ \emph {et~al.}(2023)\citenamefont {Qian}, \citenamefont {Luo}, \citenamefont {Li}, \citenamefont {Xiong}, \citenamefont {Wang}, \citenamefont {Zhang}, \citenamefont {Gang}, \citenamefont {Li}, \citenamefont {Jiang},\ and\ \citenamefont {Yang}}]{Qian2023-at}%
  \BibitemOpen
  \bibfield  {author} {\bibinfo {author} {\bibfnamefont {Y.}~\bibnamefont {Qian}}, \bibinfo {author} {\bibfnamefont {Y.}~\bibnamefont {Luo}}, \bibinfo {author} {\bibfnamefont {Y.}~\bibnamefont {Li}}, \bibinfo {author} {\bibfnamefont {T.}~\bibnamefont {Xiong}}, \bibinfo {author} {\bibfnamefont {L.}~\bibnamefont {Wang}}, \bibinfo {author} {\bibfnamefont {W.}~\bibnamefont {Zhang}}, \bibinfo {author} {\bibfnamefont {S.}~\bibnamefont {Gang}}, \bibinfo {author} {\bibfnamefont {X.}~\bibnamefont {Li}}, \bibinfo {author} {\bibfnamefont {Q.}~\bibnamefont {Jiang}},\ and\ \bibinfo {author} {\bibfnamefont {J.}~\bibnamefont {Yang}},\ }\bibfield  {title} {\bibinfo {title} {Enhanced electromagnetic wave absorption, thermal conductivity and flame retardancy of {BCN@LDH/EP} for advanced electronic packing materials},\ }\href@noop {} {\bibfield  {journal} {\bibinfo  {journal} {Chem. Eng. J.}\ }\textbf {\bibinfo {volume} {467}},\ \bibinfo {pages} {143433} (\bibinfo {year} {2023})}\BibitemShut {NoStop}%
\bibitem [{\citenamefont {Huang}\ \emph {et~al.}(2023)\citenamefont {Huang}, \citenamefont {Wang}, \citenamefont {Jin}, \citenamefont {Chen},\ and\ \citenamefont {Wu}}]{Huang2023}%
  \BibitemOpen
  \bibfield  {author} {\bibinfo {author} {\bibfnamefont {T.}~\bibnamefont {Huang}}, \bibinfo {author} {\bibfnamefont {T.}~\bibnamefont {Wang}}, \bibinfo {author} {\bibfnamefont {J.}~\bibnamefont {Jin}}, \bibinfo {author} {\bibfnamefont {M.}~\bibnamefont {Chen}},\ and\ \bibinfo {author} {\bibfnamefont {L.}~\bibnamefont {Wu}},\ }\bibfield  {title} {\bibinfo {title} {Design of silicon rubber/bn film with high through-plane thermal conductivity and ultra-low contact resistance},\ }\href {https://doi.org/10.1016/j.cej.2023.143874} {\bibfield  {journal} {\bibinfo  {journal} {Chemical Engineering Journal}\ }\textbf {\bibinfo {volume} {469}},\ \bibinfo {pages} {143874} (\bibinfo {year} {2023})}\BibitemShut {NoStop}%
\bibitem [{\citenamefont {Calderon}\ \emph {et~al.}(2015)\citenamefont {Calderon}, \citenamefont {Plata}, \citenamefont {Toher}, \citenamefont {Oses}, \citenamefont {Levy}, \citenamefont {Fornari}, \citenamefont {Natan}, \citenamefont {Mehl}, \citenamefont {Hart}, \citenamefont {Buongiorno~Nardelli},\ and\ \citenamefont {Curtarolo}}]{Calderon2015}%
  \BibitemOpen
  \bibfield  {author} {\bibinfo {author} {\bibfnamefont {C.~E.}\ \bibnamefont {Calderon}}, \bibinfo {author} {\bibfnamefont {J.~J.}\ \bibnamefont {Plata}}, \bibinfo {author} {\bibfnamefont {C.}~\bibnamefont {Toher}}, \bibinfo {author} {\bibfnamefont {C.}~\bibnamefont {Oses}}, \bibinfo {author} {\bibfnamefont {O.}~\bibnamefont {Levy}}, \bibinfo {author} {\bibfnamefont {M.}~\bibnamefont {Fornari}}, \bibinfo {author} {\bibfnamefont {A.}~\bibnamefont {Natan}}, \bibinfo {author} {\bibfnamefont {M.~J.}\ \bibnamefont {Mehl}}, \bibinfo {author} {\bibfnamefont {G.}~\bibnamefont {Hart}}, \bibinfo {author} {\bibfnamefont {M.}~\bibnamefont {Buongiorno~Nardelli}},\ and\ \bibinfo {author} {\bibfnamefont {S.}~\bibnamefont {Curtarolo}},\ }\bibfield  {title} {\bibinfo {title} {The aflow standard for high-throughput materials science calculations},\ }\href {https://doi.org/10.1016/j.commatsci.2015.07.019} {\bibfield  {journal} {\bibinfo  {journal} {Computational Materials Science}\ }\textbf {\bibinfo {volume} {108}},\ \bibinfo
  {pages} {233–238} (\bibinfo {year} {2015})}\BibitemShut {NoStop}%
\bibitem [{\citenamefont {Kirklin}\ \emph {et~al.}(2015)\citenamefont {Kirklin}, \citenamefont {Saal}, \citenamefont {Meredig}, \citenamefont {Thompson}, \citenamefont {Doak}, \citenamefont {Aykol}, \citenamefont {R\"{u}hl},\ and\ \citenamefont {Wolverton}}]{Kirklin2015}%
  \BibitemOpen
  \bibfield  {author} {\bibinfo {author} {\bibfnamefont {S.}~\bibnamefont {Kirklin}}, \bibinfo {author} {\bibfnamefont {J.~E.}\ \bibnamefont {Saal}}, \bibinfo {author} {\bibfnamefont {B.}~\bibnamefont {Meredig}}, \bibinfo {author} {\bibfnamefont {A.}~\bibnamefont {Thompson}}, \bibinfo {author} {\bibfnamefont {J.~W.}\ \bibnamefont {Doak}}, \bibinfo {author} {\bibfnamefont {M.}~\bibnamefont {Aykol}}, \bibinfo {author} {\bibfnamefont {S.}~\bibnamefont {R\"{u}hl}},\ and\ \bibinfo {author} {\bibfnamefont {C.}~\bibnamefont {Wolverton}},\ }\bibfield  {title} {\bibinfo {title} {The open quantum materials database (oqmd): assessing the accuracy of dft formation energies},\ }\bibfield  {journal} {\bibinfo  {journal} {npj Computational Materials}\ }\textbf {\bibinfo {volume} {1}},\ \href {https://doi.org/10.1038/npjcompumats.2015.10} {10.1038/npjcompumats.2015.10} (\bibinfo {year} {2015})\BibitemShut {NoStop}%
\bibitem [{\citenamefont {Jain}\ \emph {et~al.}(2013)\citenamefont {Jain}, \citenamefont {Ong}, \citenamefont {Hautier}, \citenamefont {Chen}, \citenamefont {Richards}, \citenamefont {Dacek}, \citenamefont {Cholia}, \citenamefont {Gunter}, \citenamefont {Skinner}, \citenamefont {Ceder},\ and\ \citenamefont {Persson}}]{Jain2013}%
  \BibitemOpen
  \bibfield  {author} {\bibinfo {author} {\bibfnamefont {A.}~\bibnamefont {Jain}}, \bibinfo {author} {\bibfnamefont {S.~P.}\ \bibnamefont {Ong}}, \bibinfo {author} {\bibfnamefont {G.}~\bibnamefont {Hautier}}, \bibinfo {author} {\bibfnamefont {W.}~\bibnamefont {Chen}}, \bibinfo {author} {\bibfnamefont {W.~D.}\ \bibnamefont {Richards}}, \bibinfo {author} {\bibfnamefont {S.}~\bibnamefont {Dacek}}, \bibinfo {author} {\bibfnamefont {S.}~\bibnamefont {Cholia}}, \bibinfo {author} {\bibfnamefont {D.}~\bibnamefont {Gunter}}, \bibinfo {author} {\bibfnamefont {D.}~\bibnamefont {Skinner}}, \bibinfo {author} {\bibfnamefont {G.}~\bibnamefont {Ceder}},\ and\ \bibinfo {author} {\bibfnamefont {K.~A.}\ \bibnamefont {Persson}},\ }\bibfield  {title} {\bibinfo {title} {Commentary: The materials project: A materials genome approach to accelerating materials innovation},\ }\bibfield  {journal} {\bibinfo  {journal} {APL Materials}\ }\textbf {\bibinfo {volume} {1}},\ \href {https://doi.org/10.1063/1.4812323} {10.1063/1.4812323}
  (\bibinfo {year} {2013})\BibitemShut {NoStop}%
\bibitem [{\citenamefont {Choudhary}\ \emph {et~al.}(2020)\citenamefont {Choudhary}, \citenamefont {Garrity}, \citenamefont {Reid}, \citenamefont {DeCost}, \citenamefont {Biacchi}, \citenamefont {Hight~Walker}, \citenamefont {Trautt}, \citenamefont {Hattrick-Simpers}, \citenamefont {Kusne}, \citenamefont {Centrone}, \citenamefont {Davydov}, \citenamefont {Jiang}, \citenamefont {Pachter}, \citenamefont {Cheon}, \citenamefont {Reed}, \citenamefont {Agrawal}, \citenamefont {Qian}, \citenamefont {Sharma}, \citenamefont {Zhuang}, \citenamefont {Kalinin}, \citenamefont {Sumpter}, \citenamefont {Pilania}, \citenamefont {Acar}, \citenamefont {Mandal}, \citenamefont {Haule}, \citenamefont {Vanderbilt}, \citenamefont {Rabe},\ and\ \citenamefont {Tavazza}}]{Choudhary2020}%
  \BibitemOpen
  \bibfield  {author} {\bibinfo {author} {\bibfnamefont {K.}~\bibnamefont {Choudhary}}, \bibinfo {author} {\bibfnamefont {K.~F.}\ \bibnamefont {Garrity}}, \bibinfo {author} {\bibfnamefont {A.~C.~E.}\ \bibnamefont {Reid}}, \bibinfo {author} {\bibfnamefont {B.}~\bibnamefont {DeCost}}, \bibinfo {author} {\bibfnamefont {A.~J.}\ \bibnamefont {Biacchi}}, \bibinfo {author} {\bibfnamefont {A.~R.}\ \bibnamefont {Hight~Walker}}, \bibinfo {author} {\bibfnamefont {Z.}~\bibnamefont {Trautt}}, \bibinfo {author} {\bibfnamefont {J.}~\bibnamefont {Hattrick-Simpers}}, \bibinfo {author} {\bibfnamefont {A.~G.}\ \bibnamefont {Kusne}}, \bibinfo {author} {\bibfnamefont {A.}~\bibnamefont {Centrone}}, \bibinfo {author} {\bibfnamefont {A.}~\bibnamefont {Davydov}}, \bibinfo {author} {\bibfnamefont {J.}~\bibnamefont {Jiang}}, \bibinfo {author} {\bibfnamefont {R.}~\bibnamefont {Pachter}}, \bibinfo {author} {\bibfnamefont {G.}~\bibnamefont {Cheon}}, \bibinfo {author} {\bibfnamefont {E.}~\bibnamefont {Reed}}, \bibinfo {author}
  {\bibfnamefont {A.}~\bibnamefont {Agrawal}}, \bibinfo {author} {\bibfnamefont {X.}~\bibnamefont {Qian}}, \bibinfo {author} {\bibfnamefont {V.}~\bibnamefont {Sharma}}, \bibinfo {author} {\bibfnamefont {H.}~\bibnamefont {Zhuang}}, \bibinfo {author} {\bibfnamefont {S.~V.}\ \bibnamefont {Kalinin}}, \bibinfo {author} {\bibfnamefont {B.~G.}\ \bibnamefont {Sumpter}}, \bibinfo {author} {\bibfnamefont {G.}~\bibnamefont {Pilania}}, \bibinfo {author} {\bibfnamefont {P.}~\bibnamefont {Acar}}, \bibinfo {author} {\bibfnamefont {S.}~\bibnamefont {Mandal}}, \bibinfo {author} {\bibfnamefont {K.}~\bibnamefont {Haule}}, \bibinfo {author} {\bibfnamefont {D.}~\bibnamefont {Vanderbilt}}, \bibinfo {author} {\bibfnamefont {K.}~\bibnamefont {Rabe}},\ and\ \bibinfo {author} {\bibfnamefont {F.}~\bibnamefont {Tavazza}},\ }\bibfield  {title} {\bibinfo {title} {The joint automated repository for various integrated simulations (jarvis) for data-driven materials design},\ }\bibfield  {journal} {\bibinfo  {journal} {npj Computational
  Materials}\ }\textbf {\bibinfo {volume} {6}},\ \href {https://doi.org/10.1038/s41524-020-00440-1} {10.1038/s41524-020-00440-1} (\bibinfo {year} {2020})\BibitemShut {NoStop}%
\bibitem [{\citenamefont {Blanco}\ \emph {et~al.}(2004)\citenamefont {Blanco}, \citenamefont {Francisco},\ and\ \citenamefont {Luaña}}]{Blanco2004}%
  \BibitemOpen
  \bibfield  {author} {\bibinfo {author} {\bibfnamefont {M.}~\bibnamefont {Blanco}}, \bibinfo {author} {\bibfnamefont {E.}~\bibnamefont {Francisco}},\ and\ \bibinfo {author} {\bibfnamefont {V.}~\bibnamefont {Luaña}},\ }\bibfield  {title} {\bibinfo {title} {Gibbs: isothermal-isobaric thermodynamics of solids from energy curves using a quasi-harmonic debye model},\ }\href {https://doi.org/10.1016/j.comphy.2003.12.001} {\bibfield  {journal} {\bibinfo  {journal} {Computer Physics Communications}\ }\textbf {\bibinfo {volume} {158}},\ \bibinfo {pages} {57–72} (\bibinfo {year} {2004})}\BibitemShut {NoStop}%
\bibitem [{\citenamefont {Toher}\ \emph {et~al.}(2014)\citenamefont {Toher}, \citenamefont {Plata}, \citenamefont {Levy}, \citenamefont {de~Jong}, \citenamefont {Asta}, \citenamefont {Nardelli},\ and\ \citenamefont {Curtarolo}}]{Toher2014}%
  \BibitemOpen
  \bibfield  {author} {\bibinfo {author} {\bibfnamefont {C.}~\bibnamefont {Toher}}, \bibinfo {author} {\bibfnamefont {J.~J.}\ \bibnamefont {Plata}}, \bibinfo {author} {\bibfnamefont {O.}~\bibnamefont {Levy}}, \bibinfo {author} {\bibfnamefont {M.}~\bibnamefont {de~Jong}}, \bibinfo {author} {\bibfnamefont {M.}~\bibnamefont {Asta}}, \bibinfo {author} {\bibfnamefont {M.~B.}\ \bibnamefont {Nardelli}},\ and\ \bibinfo {author} {\bibfnamefont {S.}~\bibnamefont {Curtarolo}},\ }\bibfield  {title} {\bibinfo {title} {High-throughput computational screening of thermal conductivity, debye temperature, and gr\"{u}neisen parameter using a quasiharmonic debye model},\ }\bibfield  {journal} {\bibinfo  {journal} {Physical Review B}\ }\textbf {\bibinfo {volume} {90}},\ \href {https://doi.org/10.1103/physrevb.90.174107} {10.1103/physrevb.90.174107} (\bibinfo {year} {2014})\BibitemShut {NoStop}%
\bibitem [{\citenamefont {Zhuang}\ \emph {et~al.}(2019)\citenamefont {Zhuang}, \citenamefont {Qi}, \citenamefont {Duan}, \citenamefont {Xi}, \citenamefont {Zhu}, \citenamefont {Zhu}, \citenamefont {Xiong},\ and\ \citenamefont {He}}]{surveyTC}%
  \BibitemOpen
  \bibfield  {author} {\bibinfo {author} {\bibfnamefont {F.}~\bibnamefont {Zhuang}}, \bibinfo {author} {\bibfnamefont {Z.}~\bibnamefont {Qi}}, \bibinfo {author} {\bibfnamefont {K.}~\bibnamefont {Duan}}, \bibinfo {author} {\bibfnamefont {D.}~\bibnamefont {Xi}}, \bibinfo {author} {\bibfnamefont {Y.}~\bibnamefont {Zhu}}, \bibinfo {author} {\bibfnamefont {H.}~\bibnamefont {Zhu}}, \bibinfo {author} {\bibfnamefont {H.}~\bibnamefont {Xiong}},\ and\ \bibinfo {author} {\bibfnamefont {Q.}~\bibnamefont {He}},\ }\href {https://doi.org/10.48550/ARXIV.1911.02685} {\bibinfo {title} {A comprehensive survey on transfer learning}} (\bibinfo {year} {2019})\BibitemShut {NoStop}%
\bibitem [{\citenamefont {Kong}\ \emph {et~al.}(2021)\citenamefont {Kong}, \citenamefont {Guevarra}, \citenamefont {Gomes},\ and\ \citenamefont {Gregoire}}]{Kong2021}%
  \BibitemOpen
  \bibfield  {author} {\bibinfo {author} {\bibfnamefont {S.}~\bibnamefont {Kong}}, \bibinfo {author} {\bibfnamefont {D.}~\bibnamefont {Guevarra}}, \bibinfo {author} {\bibfnamefont {C.~P.}\ \bibnamefont {Gomes}},\ and\ \bibinfo {author} {\bibfnamefont {J.~M.}\ \bibnamefont {Gregoire}},\ }\bibfield  {title} {\bibinfo {title} {Materials representation and transfer learning for multi-property prediction},\ }\bibfield  {journal} {\bibinfo  {journal} {Applied Physics Reviews}\ }\textbf {\bibinfo {volume} {8}},\ \href {https://doi.org/10.1063/5.0047066} {10.1063/5.0047066} (\bibinfo {year} {2021})\BibitemShut {NoStop}%
\bibitem [{\citenamefont {Hoffmann}\ \emph {et~al.}(2023)\citenamefont {Hoffmann}, \citenamefont {Schmidt}, \citenamefont {Botti},\ and\ \citenamefont {Marques}}]{surveyTC2}%
  \BibitemOpen
  \bibfield  {author} {\bibinfo {author} {\bibfnamefont {N.}~\bibnamefont {Hoffmann}}, \bibinfo {author} {\bibfnamefont {J.}~\bibnamefont {Schmidt}}, \bibinfo {author} {\bibfnamefont {S.}~\bibnamefont {Botti}},\ and\ \bibinfo {author} {\bibfnamefont {M.~A.~L.}\ \bibnamefont {Marques}},\ }\href {https://doi.org/10.48550/ARXIV.2303.03000} {\bibinfo {title} {Transfer learning on large datasets for the accurate prediction of material properties}} (\bibinfo {year} {2023})\BibitemShut {NoStop}%
\bibitem [{\citenamefont {Seko}\ \emph {et~al.}(2015)\citenamefont {Seko}, \citenamefont {Togo}, \citenamefont {Hayashi}, \citenamefont {Tsuda}, \citenamefont {Chaput},\ and\ \citenamefont {Tanaka}}]{Seko2015}%
  \BibitemOpen
  \bibfield  {author} {\bibinfo {author} {\bibfnamefont {A.}~\bibnamefont {Seko}}, \bibinfo {author} {\bibfnamefont {A.}~\bibnamefont {Togo}}, \bibinfo {author} {\bibfnamefont {H.}~\bibnamefont {Hayashi}}, \bibinfo {author} {\bibfnamefont {K.}~\bibnamefont {Tsuda}}, \bibinfo {author} {\bibfnamefont {L.}~\bibnamefont {Chaput}},\ and\ \bibinfo {author} {\bibfnamefont {I.}~\bibnamefont {Tanaka}},\ }\bibfield  {title} {\bibinfo {title} {Prediction of low-thermal-conductivity compounds with first-principles anharmonic lattice-dynamics calculations and bayesian optimization},\ }\bibfield  {journal} {\bibinfo  {journal} {Physical Review Letters}\ }\textbf {\bibinfo {volume} {115}},\ \href {https://doi.org/10.1103/physrevlett.115.205901} {10.1103/physrevlett.115.205901} (\bibinfo {year} {2015})\BibitemShut {NoStop}%
\bibitem [{\citenamefont {Togo}\ \emph {et~al.}(2015)\citenamefont {Togo}, \citenamefont {Chaput},\ and\ \citenamefont {Tanaka}}]{Togo2015}%
  \BibitemOpen
  \bibfield  {author} {\bibinfo {author} {\bibfnamefont {A.}~\bibnamefont {Togo}}, \bibinfo {author} {\bibfnamefont {L.}~\bibnamefont {Chaput}},\ and\ \bibinfo {author} {\bibfnamefont {I.}~\bibnamefont {Tanaka}},\ }\bibfield  {title} {\bibinfo {title} {Distributions of phonon lifetimes in brillouin zones},\ }\bibfield  {journal} {\bibinfo  {journal} {Physical Review B}\ }\textbf {\bibinfo {volume} {91}},\ \href {https://doi.org/10.1103/physrevb.91.094306} {10.1103/physrevb.91.094306} (\bibinfo {year} {2015})\BibitemShut {NoStop}%
\bibitem [{\citenamefont {Togo}\ \emph {et~al.}(2023)\citenamefont {Togo}, \citenamefont {Chaput}, \citenamefont {Tadano},\ and\ \citenamefont {Tanaka}}]{phonopy-phono3py-JPCM}%
  \BibitemOpen
  \bibfield  {author} {\bibinfo {author} {\bibfnamefont {A.}~\bibnamefont {Togo}}, \bibinfo {author} {\bibfnamefont {L.}~\bibnamefont {Chaput}}, \bibinfo {author} {\bibfnamefont {T.}~\bibnamefont {Tadano}},\ and\ \bibinfo {author} {\bibfnamefont {I.}~\bibnamefont {Tanaka}},\ }\bibfield  {title} {\bibinfo {title} {Implementation strategies in phonopy and phono3py},\ }\href {https://doi.org/10.1088/1361-648X/acd831} {\bibfield  {journal} {\bibinfo  {journal} {J. Phys. Condens. Matter}\ }\textbf {\bibinfo {volume} {35}},\ \bibinfo {pages} {353001} (\bibinfo {year} {2023})}\BibitemShut {NoStop}%
\bibitem [{\citenamefont {Chaput}(2013)}]{PhysRevLett.110.265506}%
  \BibitemOpen
  \bibfield  {author} {\bibinfo {author} {\bibfnamefont {L.}~\bibnamefont {Chaput}},\ }\bibfield  {title} {\bibinfo {title} {Direct solution to the linearized phonon boltzmann equation},\ }\href {https://doi.org/10.1103/PhysRevLett.110.265506} {\bibfield  {journal} {\bibinfo  {journal} {Phys. Rev. Lett.}\ }\textbf {\bibinfo {volume} {110}},\ \bibinfo {pages} {265506} (\bibinfo {year} {2013})}\BibitemShut {NoStop}%
\bibitem [{\citenamefont {Miyazaki}\ \emph {et~al.}(2021)\citenamefont {Miyazaki}, \citenamefont {Tamura}, \citenamefont {Mikami}, \citenamefont {Watanabe}, \citenamefont {Ide}, \citenamefont {Ozkendir},\ and\ \citenamefont {Nishino}}]{Miyazaki2021}%
  \BibitemOpen
  \bibfield  {author} {\bibinfo {author} {\bibfnamefont {H.}~\bibnamefont {Miyazaki}}, \bibinfo {author} {\bibfnamefont {T.}~\bibnamefont {Tamura}}, \bibinfo {author} {\bibfnamefont {M.}~\bibnamefont {Mikami}}, \bibinfo {author} {\bibfnamefont {K.}~\bibnamefont {Watanabe}}, \bibinfo {author} {\bibfnamefont {N.}~\bibnamefont {Ide}}, \bibinfo {author} {\bibfnamefont {O.~M.}\ \bibnamefont {Ozkendir}},\ and\ \bibinfo {author} {\bibfnamefont {Y.}~\bibnamefont {Nishino}},\ }\bibfield  {title} {\bibinfo {title} {Machine learning based prediction of lattice thermal conductivity for half-heusler compounds using atomic information},\ }\bibfield  {journal} {\bibinfo  {journal} {Scientific Reports}\ }\textbf {\bibinfo {volume} {11}},\ \href {https://doi.org/10.1038/s41598-021-92030-4} {10.1038/s41598-021-92030-4} (\bibinfo {year} {2021})\BibitemShut {NoStop}%
\bibitem [{\citenamefont {Dunn}\ \emph {et~al.}(2020)\citenamefont {Dunn}, \citenamefont {Wang}, \citenamefont {Ganose}, \citenamefont {Dopp},\ and\ \citenamefont {Jain}}]{Dunn2020}%
  \BibitemOpen
  \bibfield  {author} {\bibinfo {author} {\bibfnamefont {A.}~\bibnamefont {Dunn}}, \bibinfo {author} {\bibfnamefont {Q.}~\bibnamefont {Wang}}, \bibinfo {author} {\bibfnamefont {A.}~\bibnamefont {Ganose}}, \bibinfo {author} {\bibfnamefont {D.}~\bibnamefont {Dopp}},\ and\ \bibinfo {author} {\bibfnamefont {A.}~\bibnamefont {Jain}},\ }\bibfield  {title} {\bibinfo {title} {Benchmarking materials property prediction methods: the matbench test set and automatminer reference algorithm},\ }\bibfield  {journal} {\bibinfo  {journal} {npj Computational Materials}\ }\textbf {\bibinfo {volume} {6}},\ \href {https://doi.org/10.1038/s41524-020-00406-3} {10.1038/s41524-020-00406-3} (\bibinfo {year} {2020})\BibitemShut {NoStop}%
\bibitem [{\citenamefont {Wang}\ \emph {et~al.}(2021)\citenamefont {Wang}, \citenamefont {Kauwe}, \citenamefont {Murdock},\ and\ \citenamefont {Sparks}}]{Wang2021crabnet}%
  \BibitemOpen
  \bibfield  {author} {\bibinfo {author} {\bibfnamefont {A.~Y.-T.}\ \bibnamefont {Wang}}, \bibinfo {author} {\bibfnamefont {S.~K.}\ \bibnamefont {Kauwe}}, \bibinfo {author} {\bibfnamefont {R.~J.}\ \bibnamefont {Murdock}},\ and\ \bibinfo {author} {\bibfnamefont {T.~D.}\ \bibnamefont {Sparks}},\ }\bibfield  {title} {\bibinfo {title} {Compositionally restricted attention-based network for materials property predictions},\ }\href {https://doi.org/10.1038/s41524-021-00545-1} {\bibfield  {journal} {\bibinfo  {journal} {npj Computational Materials}\ }\textbf {\bibinfo {volume} {7}},\ \bibinfo {pages} {77} (\bibinfo {year} {2021})}\BibitemShut {NoStop}%
\end{thebibliography}%

\end{document}